%% file: main.tex
\documentclass[runningheads]{llncs}

\usepackage{eccv}

\usepackage{eccvabbrv}

\usepackage{graphicx}
\usepackage{booktabs}
\usepackage{multirow}
\usepackage{xspace}
\usepackage{wrapfig}
\newcommand{\nn}{SVCR}
\usepackage{colortbl}
\newcolumntype{g}{>{\columncolor[gray]{0.9}}c}
\usepackage[accsupp]{axessibility}  

\usepackage{hyperref}

\usepackage{orcidlink}

\begin{document}

\title{Learning Structured Visual Compositional Representations for Weakly Supervised Referring Expression Comprehension}

\titlerunning{Learning Structured Visual Compositional Representations for WREC}

\author{Lian Xu\inst{1}\orcidlink{0000-0002-1759-2941} \and
Mohammed Bennamoun\inst{1}\orcidlink{0000-0002-6603-3257
} \and
Farid Boussaid\inst{1}\orcidlink{0000-0001-7250-7407} 
\and \\
Hamid Laga\inst{2}\orcidlink{0000-0002-4758-7510
}
\and
Yulan Guo\inst{3}\orcidlink{0000-0001-7051-841X}
\and
Dan Xu\inst{4}\orcidlink{0000-0003-0136-9603}
}

\authorrunning{L.~Xu et al.}

\institute{$^{1}$The University of Western Australia, 
$^{2}$Murdoch University, 
$^{3}$Sun Yat-sen University, 
$^{4}$The Hong Kong University of Science and Technology \\
\email{\{lian.xu, mohammed.bennamoun, farid.boussaid\}@uwa.edu.au, H.Laga@murdoch.edu.au, guoyulan@sysu.edu.cn, danxu@cse.ust.hk}
}

\maketitle

\begin{abstract}
Referring expression comprehension (REC) aims to localize the object in an image described by natural language. In Weakly supervised REC (WREC), existing approaches primarily operate on anchor-level visual representations. Even when enriched with auxiliary cues, relational interactions remain implicitly encoded within individual anchor features. The resulting visual representation remains flat and unary-only, limiting its ability to align with the structured nature of language. In this work, we propose a Structured Visual Compositional Representation (\nn) learning framework for WREC. Rather than implicitly encoding relations within unary anchors, the proposed \nn~explicitly models both unary object embeddings and pairwise relational embeddings, forming a structured visual representation space. We further introduce a compositional alignment mechanism that matches unary and pairwise visual representations with their corresponding textual embeddings in a unified manner, enabling compositional visual–textual matching under weak supervision. 
Extensive experiments on RefCOCO, RefCOCO+, and RefCOCOg show that the proposed \nn~achieves state-of-the-art performance. These results demonstrate the effectiveness of explicit structured visual representations and visual-textual alignment for WREC.\footnote{https://github.com/quiet5daxis/SVCR}
\end{abstract}

\section{Introduction}
\label{sec:intro}

\begin{figure*}[t]
    \centering
    \includegraphics[width=\linewidth]{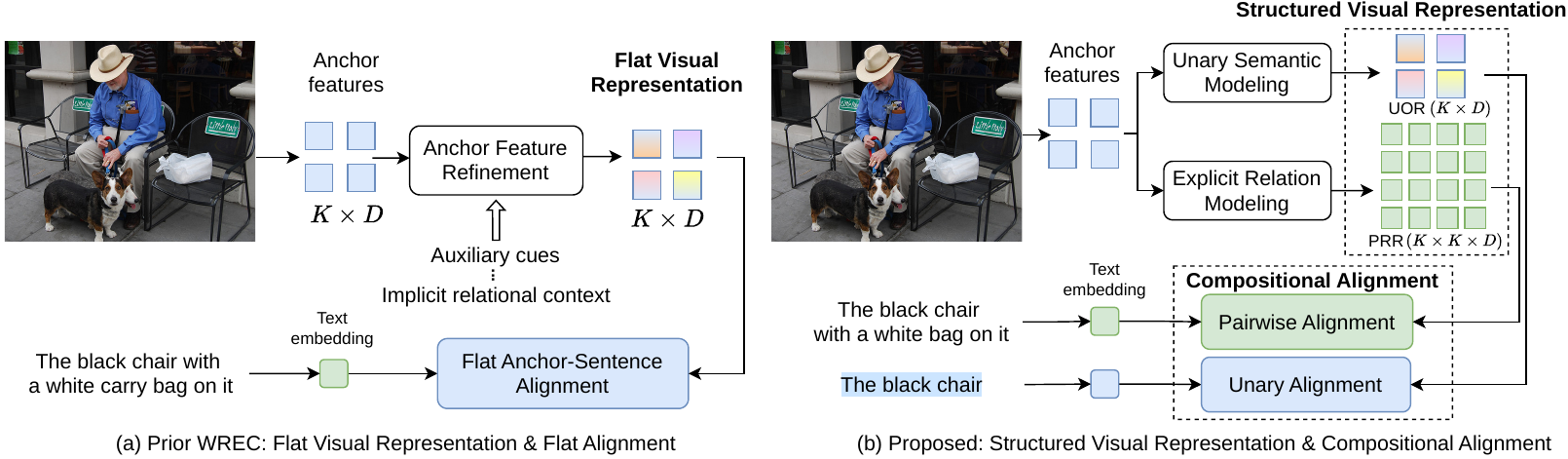}
    \caption{
    Comparison between prior WREC methods~\cite{jin2023refclip,luo2024apl,chen2025dvin} and the proposed method.
(\textbf{a}) Prior WREC approaches operate in a flat visual representation space, where auxiliary cues or relational context are aggregated into unary anchor representations ($K\times D$, with $K$ anchors and feature dimension $D$). The full expression is directly matched with these flattened representations. In contrast, (\textbf{b}) our method constructs a structured visual representation space consisting of unary object representations (UOR, $K\times D$) and pairwise relational representations (PRR, $K\times K\times D$). Based on this, we introduce compositional alignment, aligning subject-level cues with unary representations and sentence-level semantics with relational representations under weak supervision.
    }
    \label{fig:teaser}
    \vspace{-2em}
\end{figure*}

Referring Expression Comprehension (REC), also known as visual grounding, aims to localize in an image the object that corresponds to a given natural language description~\cite{yang2019cross}. 
This task is central to many real-world applications such as human-robot interaction and visual navigation. While supervised REC methods~\cite{luo2020multi, yang2020graph} have achieved remarkable performance, they require dense bounding-box annotations linking objects to referring expressions. These, however, are costly to collect at scale and limit adaptability across domains. To alleviate annotation burdens, recent works~\cite{jin2023refclip} have shifted towards weakly-supervised REC (WREC), which leverages only paired image–text data for supervision.

Unlike supervised REC, which benefits from explicit box-level supervision to directly optimize fused multimodal features, WREC predominantly relies on cross-modal contrastive alignment to associate textual descriptions with candidate regions. Recent advances~\cite{jin2023refclip,luo2024apl,chen2025dvin} in WREC have focused on one-stage anchor-based frameworks, which reformulate the task as anchor-text matching optimized through contrastive learning. These methods iteratively select top-ranked anchor-text pairs as pseudo-positives and update model parameters by reinforcing these matches via contrastive loss. They effectively perform latent assignment over region proposals in an Expectation-Maximization~\cite{dempster1977maximum} (EM)-like optimization process, which progressively refines regional vision-language alignment without box-level annotations.

Prior work has often attributed performance gaps in WREC to the coarse or category-level semantics of anchor features, which lack instance-specific attributes and fine-grained distinctions necessary for matching complex linguistic descriptions~\cite{chen2025dvin}. However, we argue that this limitation is more fundamental: \textit{both visual representations and cross-modal alignment are defined solely at the unary anchor level}.
Even when the anchors are enriched with contextual or relational cues through foundation features~\cite{chen2025dvin,cheng2025weakmcn}, handcrafted priors (\eg, spatial coordinates, category, or color)~\cite{luo2024apl}, or relational message passing~\cite{liu2021relation}, inter-object interactions are ultimately absorbed back into individual anchor embeddings, as illustrated in Fig.~\ref{fig:teaser} (a). As a result, the visual representation space remains structurally flat. However, referring expressions are inherently compositional and relational. They describe target objects not only through unary properties (e.g., ``small'', ``wooden'') but also through pairwise relations (e.g., ``on the table'', ``to the left of the man''). 
For example, in Fig.~\ref{fig:teaser}, the expression “the black chair with a white carry bag on it” requires identifying the chair through its relation to another object (the bag). When all contextual and relational signals are flattened into unary anchor embeddings, such inter-object dependencies cannot be explicitly represented. Consequently, relational semantics become entangled with appearance or category cues within unary embeddings, limiting the model’s ability to reliably ground compositional referring expressions.

To address this representation mismatch, we propose Structured Visual Compositional Representations (\nn) for WREC. Instead of encoding relational cues as refinements of unary anchors, as illustrated in Fig.~\ref{fig:teaser} (b), the proposed \nn~explicitly models two complementary components: (\textbf{i}) \textbf{Unary object embeddings}, obtained through language-guided refinement that captures instance-specific semantics. By leveraging the inherently rich and descriptive nature of language, this refinement provides a flexible supervisory signal without relying on manually engineered attribute categories. (\textbf{ii}) \textbf{Pairwise relational embeddings}, constructed by explicitly modeling interactions between anchor pairs. Rather than aggregating contextual cues back into individual anchors, we form structured pairwise representations that encode both semantic and geometric dependencies, preserving inter-object relational structure. Together, these unary and pairwise components establish a structured visual representation space that mirrors the compositional structure of referring expressions.
Based on this, we introduce a \textbf{compositional alignment} mechanism that consistently aligns unary and pairwise visual representations with their corresponding textual embeddings under weak supervision. By generalizing anchor-based matching to structured multi-granular alignment, the proposed \nn~enables compositional visual–textual matching without requiring box-level annotations.

Our main contributions are summarized as follows: 
\vspace{-0.5em}
\begin{itemize}
    \item \textbf{Structured visual representations for WREC.} 
    We introduce a structured visual representation framework that explicitly models both unary object embeddings and pairwise relational embeddings. This shifts WREC from unary-only anchor representations to structured visual modeling, addressing the fundamental representation granularity limitation in prior work.

    \item \textbf{Compositional alignment under weak supervision.} 
    We propose a compositional alignment mechanism that jointly aligns unary and pairwise visual representations with their corresponding textual embeddings, enabling structured visual–textual matching without box-level annotations.

    \item \textbf{State-of-the-art performance.} 
    Extensive experiments on RefCOCO, RefCOCO+, and RefCOCOg show that our method achieves state-of-the-art performance, validating the effectiveness of SVCR for WREC.
\end{itemize}

\section{Related Work}
\label{sec:related}

\medskip
\noindent\textbf{Weakly Supervised REC (WREC).} 
Fully supervised REC methods~\cite{yang2019dynamic, liu2021relation, liu2017referring, wang2019neighbourhood, yu2018mattnet,deng2021transvg, ma2024groma, zhang2024omg, qiu2025spatial} have achieved strong performance by learning from box- or pixel-level annotations that explicitly associate expressions with target regions. However, such dense supervision is expensive and difficult to scale, motivating the development of WREC, where models learn to ground expressions using only image–text pairs without explicit region annotations.
Early works~\cite{datta2019align2ground, akbari2019multi} formulated WREC under a Multiple Instance Learning (MIL) framework, aggregating region proposals into bag-level representations and optimizing bag–text alignment. Subsequent methods~\cite{gupta2020contrastive, zhang2020counterfactual} shifted toward direct region–text contrastive learning, often adopting an EM-like optimization process to progressively refine pseudo-positive region assignments. This line of work led to efficient one-stage anchor–text frameworks~\cite{jin2023refclip}, which reformulate WREC as anchor-level matching under contrastive supervision. To address the limited expressiveness of raw anchors extracted from pretrained detectors, recent approaches focus on anchor feature enrichment. APL~\cite{luo2024apl} incorporates auxiliary cues such as spatial coordinates and category labels into anchor embeddings. DViN~\cite{chen2025dvin} fuses multi-source foundation model features to enhance semantic diversity. WeakMCN~\cite{cheng2025weakmcn} jointly optimizes WREC and WRES to improve cross-task consistency. 
Despite these improvements, visual representations and cross-modal alignment remain defined at the unary anchor level, limiting the ability to capture structured inter-object dependencies expressed in natural language.

\noindent\textbf{Language Structure for Visual Grounding.}
Prior work has explored incorporating linguistic structure into visual grounding or other related vision-language tasks. Their use can be broadly categorized into the following:
(\textbf{i}) \textit{Attention regularization.}
Early weakly supervised phrase grounding (WSPG) methods~\cite{xiao2017weakly} employed constituency parse trees to impose structural constraints over spatial attention maps, encouraging consistency between parent and child phrases. More recent works have applied similar ideas to weakly supervised referring expression segmentation (WRES). For example, PCNet~\cite{yangboosting} decomposes the expression into shorter phrases to progressively refine visual representations and enforce consistency across their corresponding visual–textual similarity maps, while Lee~\etal~\cite{lee2023weakly} extract the noun chunks from the expression to regularize the text-to-image attention maps. None of them provide structured visual representations under weak supervision
\textbf{(ii)} \textit{Relation Modeling within Unary Representations.}
Relation modeling is widely used in fully supervised REC, where expressions are decomposed into semantic components (\eg, attributes and relations) and graph modules are employed to propagate contextual information between object proposals~\cite{yang2019dynamic, yang2020graph, liu2021relation, wang2019neighbourhood, yu2018mattnet}. A few works extend this idea to weakly supervised settings, such as ~\cite{liu2021relation}, which incorporates relational context through proposal interaction. Despite these efforts, relational cues are typically used to augment proposal features via message passing, after which the enriched proposals are still treated as unary embeddings for proposal–text matching. While this design works well in supervised REC with explicit box annotations, its role in WREC is less clear since WREC relies on contrastive anchor–text matching with latent assignment, excessive proposal interaction may reduce anchor discriminability and weaken contrastive alignment. Consequently, the final visual representation space remains structurally flat.
(\textbf{iii})~\textit{Structured representation matching.} 
Another line of work constructs structured representations for both modalities and performs graph-to-graph matching between visual scene graphs and linguistic structures~\cite{teney2017graph, lyu2020vtgraphnet}. 
Related ideas also appear in scene graph modeling works~\cite{xu2017scene, yang2018graph, zellers2018neural}. 
However, such approaches typically rely on object-level annotations or scene graph supervision to learn relational structures, making them difficult to apply in weakly supervised settings.

In contrast, we propose structured visual representations that explicitly encodes both unary objects and pairwise relations. This representation enables multi-granular vision–language alignment under weak supervision, addressing the limitations of unary-only grounding paradigms.

\begin{figure*}[t]
    \centering
    \includegraphics[width=0.95\linewidth]{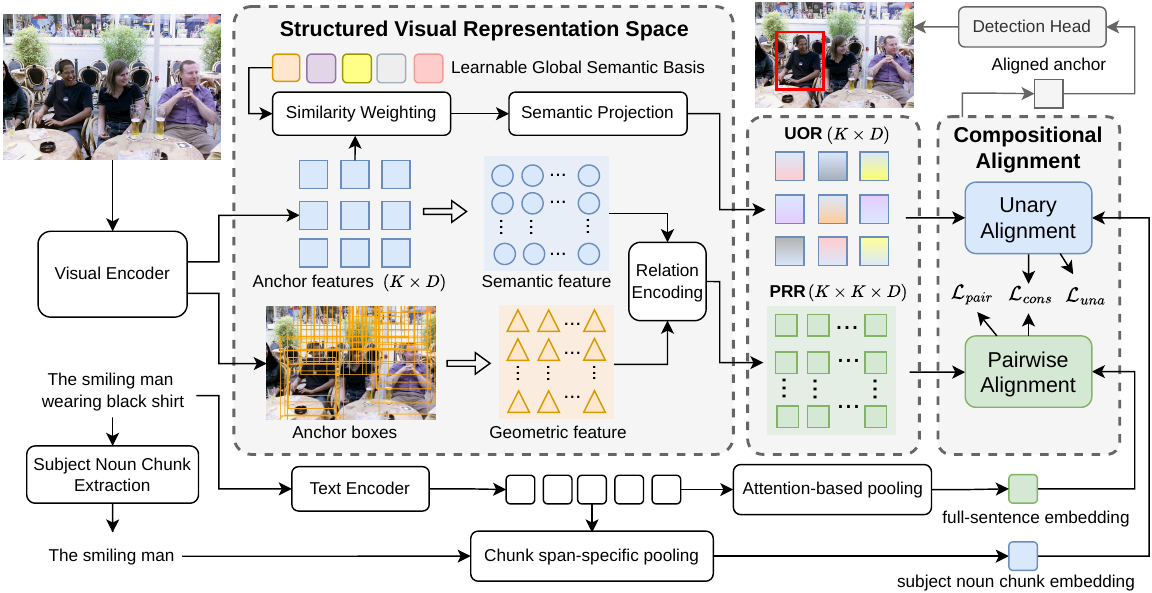}
    \vspace{-1em}
    \caption{Overview of the proposed framework. Given an input image and referring expression, we first construct a Structured Visual Representation Space from anchor features extracted by a pretrained detector. This space decomposes visual semantics into Unary Object Representations (UOR, $K \times D$) enhanced by learnable global semantic bases, and Pairwise Relational Representations (PRR, $K \times K \times D$) that encode semantic and geometric interactions among anchors, where $K$ denotes the number of anchors and $D$ the feature dimension. Based on this structured space, we introduce a compositional alignment module consisting of Unary Alignment which matches unary object representations with subject noun chunk embeddings, and Pairwise Alignment which associates relational representations with full-sentence embeddings. Training is driven by contrastive objectives at both granularities together ($\mathcal{L}_{una}$ and $\mathcal{L}_{pair}$) with a hierarchical consistency constraint ($\mathcal{L}_{cons}$), an a diversity regularization stabilizes the learned semantic bases (see Sec.~\ref{sec:svr} for details). During inference, the aligned anchor is selected and decoded by the detection head to produce the final grounding result.
    \label{overview}
    }
    \label{fig:overview}
    \vspace{-1em}
\end{figure*}

\section{Method}
\noindent\textbf{Overview.} As illustrated in Fig.~\ref{overview}, our framework follows the standard anchor-based WREC pipeline while fundamentally redefining the visual representation space for structured grounding.
Given an image–expression pair, we first encode the visual and textual modalities. A pretrained detection backbone extracts anchor-level visual features, while a lightweight language encoder produces contextualized textual embeddings that capture both object-centric and relational cues.
To bridge the structural mismatch between compositional language and unary anchor representations, we explicitly construct a \emph{structured visual representation space} composed of two complementary components: (\textbf{i}) unary object representations that enhance instance-level discriminability at the anchor level, and (\textbf{ii}) pairwise relational representations that explicitly model inter-object interactions.
Unlike prior approaches that aggregate contextual information into individual anchors, our design preserves object-level and relational-level semantics as distinct representation units.
Based on this, we perform compositional alignment, where unary and relational visual representations are matched with corresponding textual representations at different semantic granularities.

During inference, grounding is determined by jointly considering unary object compatibility and relational contextual evidence.
The anchor with the highest aggregated compositional confidence is selected, and its bounding box is decoded by the detection head. Through this design, the proposed framework moves beyond flat anchor-based matching toward structured visual–textual alignment, enabling robust grounding for both attribute-centric expressions and complex relational descriptions.

\subsection{Feature Extraction Backbone}
We use two encoders to extract visual and textual representations: a frozen YOLOv3 backbone for anchor-level visual features and a lightweight recurrent–attention encoder for textual embeddings. We further enrich the anchor features with auxiliary visual cues from visual foundation models through a dynamic routing mechanism.

\noindent\textbf{Visual Encoder.}
Given an input RGB image $\textbf{I} \in \mathbb{R}^{3\times H \times W}$, we adopt the pre-trained DarkNet-53 backbone from YOLOv3 and take anchor-level features from the detection head, which provides multi-scale anchor representations. Following prior work~\cite{jin2023refclip}, we select the top $K$ anchors with the highest objectness scores and denote their features as
$\mathbf{A} = \{\mathbf{a}_1, \mathbf{a}_2, \ldots, \mathbf{a}_K\}$, where $\mathbf{a}_i \in \mathbb{R}^d$. Motivated by prior works~\cite{chen2025dvin,cheng2025weakmcn}, we incorporate additional visual features $\mathbf{F} = \{\mathbf{f}_1, \ldots, \mathbf{f}_M\}$, where $\mathbf{f}_j \in \mathbb{R}^d$, from foundation models such as DINO v2~\cite{oquab2024dinov2} and Depth Anything v2~\cite{yangdepth} to enhance the expressiveness of $\mathbf{A}$. 
Each anchor $\mathbf{a}_i$ is enriched by dynamically routing information from $\mathbf{F}$:
\begin{equation}
\tilde{\mathbf{a}}_i = \mathbf{a}_i + \sum_{j=1}^M \alpha_{ij}\,\mathbf{f}_j, \quad 
\alpha_{ij} = 
\frac{\exp\big((\mathbf{w}_j^\top \mathbf{a}_i)\big)}
{\sum_{j'=1}^M \exp\big((\mathbf{w}_{j'}^\top \mathbf{a}_i)\big)},
\end{equation}
where $\alpha_{ij}$ denotes the routing weights, and $\mathbf{w}_j \in \mathbb{R}^d$ is a learnable projection vector associated with the $j$-th foundation feature.
The resulting enriched anchors are denoted as $\tilde{\mathbf{A}} = \{\tilde{\mathbf{a}}_1, \ldots, \tilde{\mathbf{a}}_K\}.$

\noindent\textbf{Textual Encoder.} Given a referring expression, \textit{e.g., ``the young man standing by the table”}, we first tokenize it and map tokens into embeddings $\mathbf{E} = \{\mathbf{e}_1, \ldots, \mathbf{e}_L\}$, where $\mathbf{e}_t \in \mathbb{R}^{d_w}$.
These embeddings are passed through an LSTM to obtain contextualized representations:
\begin{equation}
\mathbf{h}_t = \mathrm{LSTM}(\mathbf{e}_t, \mathbf{h}_{t-1}), 
\quad t = 1,\ldots,L.
\end{equation}

\noindent To capture higher-order dependencies, we apply several layers of self-attention:
\begin{equation}
\mathbf{z}_t = \mathrm{SelfAttn}(\mathbf{h}_t, \mathbf{H}), 
\quad \mathbf{H} = \{\mathbf{h}_1,\ldots,\mathbf{h}_L\}.
\end{equation}

\noindent From the contextualized sequence $\{\mathbf{z}_t\}_{t=1}^L$, 
we obtain two textual embeddings.  First, a token-level attention pooling layer produces the global sentence embedding $\mathbf{s} \in \mathbb{R}^d$:
\begin{equation}
\beta_t = 
\frac{\exp(\mathbf{w}^\top \tanh(\mathbf{W}\mathbf{z}_t))}
{\sum_{t'} \exp(\mathbf{w}^\top \tanh(\mathbf{W}\mathbf{z}_{t'}))}, 
\quad
\mathbf{s} = \sum_{t=1}^L \beta_t \mathbf{z}_t.
\end{equation}

\noindent Second, we extract the embedding of the subject noun chunk (\textit{i.e.,``the young man”}) by pooling the token representations: $\mathbf{s}_c = 1/(e-b+1) \sum_{t=b}^{e} \mathbf{z}_t$, where $(b,e)$ corresponds to the chunk span. 

\subsection{Structured Visual Representation Space}
\label{sec:svr}
To bridge the structural gap between language and visual representations, we explicitly construct a structured visual representation space composed of unary object representations and pairwise relational representations. Unlike conventional anchor refinement approaches that encode contextual cues within individual anchor embeddings, we preserve object-level and relational-level semantics as distinct but complementary representation units.

\noindent\textbf{Unary Object Representations.}
Given the set of enriched anchor features $\tilde{\mathbf{A}} = \{\tilde{\mathbf{a}}_1, \ldots, \tilde{\mathbf{a}}_K\}$ obtained from the backbone, we we first perform \emph{unary semantic modeling} to construct unary object representations that enhance instance-level discriminability while preserving localization. 
Rather than manually specifying attribute categories, we introduce $N$ learnable semantic basis vectors
$\mathbf{P} = \{\mathbf{p}_1, \ldots, \mathbf{p}_N\}$ that span a shared semantic subspace by the entire dataset, where $\mathbf{p}_j \in \mathbb{R}^d$. Each anchor $\tilde{\mathbf{a}}_i$ is projected onto this subspace to obtain a refined semantic component:
\begin{equation}
\mathbf{a}_i^{\text{una}} = \sum_{j=1}^N \gamma_{ij}\,\mathbf{p}_j, \quad
\gamma_{ij} = \frac{\exp(\mathbf{p}_j^\top \tilde{\mathbf{a}}_i)} 
{\sum_{j'=1}^N \exp(\mathbf{p}_{j'}^\top \tilde{\mathbf{a}}_i)}.
\end{equation}
This projection allows anchors to adaptively emphasize semantically relevant directions under weak language supervision.
The refined semantic component is fused with the original anchor feature through gated integration:
\begin{equation}
\hat{\mathbf{a}}_i = \mathbf{g}_i \odot \mathbf{a}_i^{\text{una}} 
+ (1-\mathbf{g}_i) \odot \tilde{\mathbf{a}}_i,\quad 
\mathbf{g}_i = \sigma\big(\mathrm{MLP}([\tilde{\mathbf{a}}_i \,\|\, \mathbf{a}_i^{\text{una}}])\big),
\end{equation}
where $\sigma(\cdot)$ denotes the sigmoid activation and $[\,\cdot \,\|\, \cdot\,]$ represents concatenation. 
The resulting $\hat{\mathbf{A}} = \{\hat{\mathbf{a}}_1, \ldots, \hat{\mathbf{a}}_K\}$ serve as the unary object representation.  

To encourage the semantic bases to capture diverse semantic directions, we regularize them with an orthogonality constraint:
\begin{equation}
\mathcal{L}_{\text{div}} = 
\big\| \mathbf{B}\mathbf{B}^\top - \mathbf{I} \big\|_F^2,
\end{equation}
where $\mathbf{B} = \mathrm{normalize}(\mathbf{P},\ell_2)$, $\textbf{I}$ denotes the identity matrix, and $\|\cdot\|_F$ denotes the Frobenius norm. These unary representations capture instance-specific properties but remain object-centric units in the structured space.

\noindent\textbf{Pairwise Relational Representations.}
While unary semantics enhance intra-anchor discrimination, 
they do not explicitly encode inter-object dependencies essential for grounding relational expressions. 
To address this, we perform \emph{explicit relation modeling} by constructing pairwise relational representations $\mathbf{r}_{ij}$ for anchor pairs using their semantic and geometric features.

For any anchor pair $(i,j)$, we define
\begin{equation}
\mathbf{r}_{ij} = w_{ij} \cdot \mathrm{MLP}\big([\tilde{\mathbf{a}}_i \,\|\, \tilde{\mathbf{a}}_j \,\|\, \delta_{ij}]\big), \quad \delta_{ij} = \mathrm{MLP}(\mathbf{g}_{ij}),
\end{equation}
where $\mathbf{g}_{ij}$ denotes an 8-dimensional geometric descriptor encoding relative position, scale, overlap, and orientation:
\begin{equation}
[\Delta x, \Delta y, \log(w_j/w_i), \log(h_j/h_i), \mathrm{IoU}, d, \cos\theta, \sin\theta].  
\end{equation}
The learnable weight $w_{ij} = \sigma\big(\mathrm{MLP}([\tilde{\mathbf{a}}_i \,\|\, \tilde{\mathbf{a}}_j \,\|\, \delta_{ij}])\big)$ adaptively modulates each relation, suppressing noisy interactions.

Unlike relational message passing mechanisms that absorb contextual information into individual anchors, our formulation preserves each $\mathbf{r}_{ij}$ as an explicit relational representation unit. This enables structured modeling of context-dependent interactions that cannot be reduced to unary embeddings.

\subsection{Compositional Alignment}

Given the structured visual representations, we define a compositional alignment mechanism that matches visual and textual representations at corresponding semantic granularities. Unlike conventional anchor–text matching that operates solely at the object level, our alignment is defined over both unary object representations and pairwise relational representations.

\noindent\textbf{Unary Alignment.} We compute object-level compatibility between unary visual representations and subject-centric textual embedding $\mathbf{s}_c$:
\begin{equation}
s_i^{u} = \hat{\mathbf{a}}_i^\top \mathbf{s}_c .
\end{equation}
Under weak supervision, we optimize unary alignment using the InfoNCE contrastive objective over all anchors:

\begin{equation}
\mathcal{L}_{\text{una}} = 
- \log \frac{\exp(s_{i^*}^{u} / \tau)}
{\sum_{i=1}^K \exp(s_i^{u} / \tau)},
\end{equation}
where $i^*$ denotes the pseudo-positive anchor index selected based on the highest matching confidence under weak supervision, and $\tau$ is a temperature parameter.

\noindent\textbf{Pairwise Alignment.}
To capture relational grounding, we measure compatibility between pairwise visual representations $\mathbf{r}_{ij}$
and the full-sentence embedding $\mathbf{s}$, which encodes global relational structure:
\begin{equation}
s_{ij}^{r} = \mathbf{r}_{ij}^\top \mathbf{s} .
\end{equation}
The relational alignment objective is defined as:
\begin{equation}
\mathcal{L}_{\text{pair}} = 
- \log \frac{\exp(\mathbf{r}_{i^*}^\top \mathbf{s} / \tau)}
{\sum_{i=1}^{K^2} \exp(\mathbf{r}_i^\top \mathbf{s} / \tau)}.
\end{equation}

\noindent\textbf{Hierarchical Consistency.}
Since unary and relational alignments operate at different semantic granularities,
they should provide mutually compatible evidence.
We therefore introduce a hierarchical consistency constraint
that encourages relational confidence to concentrate on anchors that are also highly ranked under unary alignment.

Let $\mathbf{v}_{\text{una}} \in \mathbb{R}^{K}$ denote the unary similarity scores,
and let $\mathbf{V}_{\text{pair}} \in \mathbb{R}^{K \times K}$ denote the pairwise similarity matrix.
We aggregate relational scores to obtain per-anchor relational confidence $\mathbf{v}_{\text{pair}}$.
The consistency loss is defined as:
\begin{equation}
\mathcal{L}_{\text{cons}} = - \log m,
\quad
m = \sum_{i \in \mathrm{TopK}(\mathbf{v}_{\text{una}})} \mathbf{v}_{\text{pair}}(i).
\end{equation}

\noindent\textbf{Overall Objective.}
The final training objective is:
\begin{equation}
\mathcal{L} =
\mathcal{L}_{\text{una}}
+
\mathcal{L}_{\text{pair}}
+
\mathcal{L}_{\text{div}}
+
\mathcal{L}_{\text{cons}}.
\end{equation}

\subsection{Grounding Inference}

At inference time, we combine relational and unary alignment scores to produce the final grounding prediction.
Relational confidence for each anchor is computed based on a symmetric similarity score as:
\begin{equation}
\mathrm{Sim}^{\text{pair}}_i = \frac{1}{2} \left( \log \sum_j \exp(\mathbf{r}_{ij}^\top \mathbf{s}) + \log \sum_j \exp(\mathbf{r}_{ji}^\top \mathbf{s}) \right).
\end{equation}
Unary confidence is computed as:
\begin{equation}
\mathrm{Sim}^{\text{una}}_i = \hat{\mathbf{a}}_i^\top \mathbf{s}_{\text{c}}.
\end{equation}
The final score is obtained via weighted aggregation:
\begin{equation}
\mathrm{Score}_i = \lambda \, \cdot \mathrm{Sim}^{\text{pair}}_i + (1-\lambda)\, \cdot \mathrm{Sim}^{\text{una}}_i ,
\end{equation}
where $\lambda$ is fixed to $0.6$ for all datasets in our experiments.
The anchor with the highest fused score is selected. Its bounding box is decoded via the detection head to produce the final grounding output. See detailed efficiency analysis in the Supplementary.

\section{Experiments}

\subsection{Experimental Settings}

\noindent\textbf{Datasets.} Following the common practice~\cite{jin2023refclip,luo2024apl,chen2025dvin}, we evaluate our method on three widely used referring expression comprehension benchmarks: RefCOCO~\cite{yu2016modeling}, RefCOCO+~\cite{yu2016modeling}, and RefCOCOg~\cite{nagaraja2016modeling}, all derived from the MSCOCO dataset~\cite{lin2014microsoft} with object-level referring expressions. 
We use the standard splits for fair comparison. See more details in the Supplementary.

\noindent\textbf{Evaluation metric.} IoU@0.5 is used to measure the accuracy of the predicted bounding boxes whose Intersection over Union (IoU) with the ground-truth bounding boxes are above 0.5. 

\noindent\textbf{Implementation details.} 
Following prior work~\cite{jin2023refclip, chen2025dvin, cheng2025weakmcn}, we used the pretrained YOLOv3 as the default detection network unless otherwise specified, and set the number of candidate anchors $K$ to 17. 
Following DViN~\cite{chen2025dvin} and WeakMCN~\cite{cheng2025weakmcn}, we incorporated pretrained visual foundation models, DINOv2-base~\cite{oquab2024dinov2} model and CLIP ViT-base~\cite{radford2021learning}, to enrich anchor features. We additionally used the Depth Anythingv2-small~\cite{yangdepth} model. We initialized the text embeddings with GloVe~\cite{pennington2014glove} pretrained weights and set the maximum sequence length to 15. 
All experiments were conducted on a single NVIDIA RTX 3090 GPU. See more details in the Supplementary.

\begin{table*}[t]
\caption{Performance comparison with the state-of-the-art methods in REC and WREC on RefCOCO, RefCOCO+, and RefCOCOg. B: bounding box. T: text supervision (Sup.). $^{*}$ denotes results based on YOLOv5, while all other WREC methods are based on YOLOv3.
The best and the second best results are \textbf{bold} and \underline{underlined}.
}
\vspace{-0.5em}
\centering
\resizebox{\linewidth}{!}
{
\begin{tabular}{clcccccccc}
\toprule
\multirow{2}{*}{Task}&\multirow{2}{*}{Methods} &\multirow{2}{*}{Sup.}    & \multicolumn{3}{c}{RefCOCO}                                        & \multicolumn{3}{c}{RefCOCO+}                                       & RefCOCOg                 \\
&& & val& testA  & testB     & val& testA    & testB         & val       \\ \midrule
\multirow{4}{*}{REC} & SimVG-B (NeurIPS24)~\cite{dai2024simvg} & B &87.63& 90.22& 84.04 &78.65& 83.36 &71.82& 78.81 \\
&OneRef-B (NeurIPS24)~\cite{xiao2024oneref} &B&88.75 &90.95 &85.34 &80.43 &86.46& 74.26& 83.68 \\
&RefFormer-B (NeurIPS24)~\cite{wang2024referencing} & B &86.52& 90.24 &81.42& 76.58& 83.69& 67.38 &77.80 \\
&TCRT (CVPR2025)~\cite{chen2025task} &B & 91.07 &92.85& 86.24& 85.75 &90.76& 77.59& 86.81\\
\midrule 
\multirow{8}{*}{WREC} 
&ARN (ICCV19)~\cite{liu2019adaptive}&T&32.17 &35.25& 30.28 &32.78& 34.35& 32.13& 33.09\\
&IGN (NeurIPS 20)~\cite{zhang2020counterfactual}&T&34.78 &37.64 &32.59& 34.29& 36.91&33.56& 34.92 \\
&DTWREG (PAMI21)~\cite{sun2021discriminative}&T&38.35 &39.51& 37.01& 38.19& 39.91& 37.09 &42.54\\
&RefCLIP (CVPR23)~\cite{jin2023refclip} & T & 60.36 &58.58 &57.13& 40.39 &40.45 &38.86 &47.87 \\
&APL (ECCV24)~\cite{luo2024apl}  & T&64.51& 61.91 &63.57& 42.70 &42.84& 39.80 &50.22 \\
&WeakMCN (CVPR25)~\cite{cheng2025weakmcn}  & T &69.20 &69.88 &62.63 &51.90& 57.33& 43.10& 54.62\\
&DViN (CVPR25)~\cite{chen2025dvin}  & T& 67.67& 70.90& 59.39 &52.54& 57.52 &\underline{45.31} &55.04\\
&\nn~(Ours) & T & \underline{71.06}&\underline{72.60}&\underline{65.55} & \underline{54.17}&\underline{60.22}&44.63&\underline{56.99}\\
&\nn$^{*}$ (Ours) &T & \textbf{74.51}&	\textbf{76.93}	&\textbf{70.46}&	\textbf{56.44}&	\textbf{64.15}	&\textbf{48.68}&	\textbf{60.40} \\
\bottomrule
\end{tabular}}

\label{tab:sota}
\vspace{-1em}
\end{table*}

\subsection{Comparison with State-of-the-Art}
Table~\ref{tab:sota} reports comparisons with state-of-the-art methods under both fully supervised REC and WREC. While fully supervised methods achieve the highest scores, thanks to box-level annotations, the proposed \nn, trained without such supervision, shows competitive results (\eg, 74.51\%, 76.93\%, and 70.46\% on RefCOCO \textit{val}, \textit{testA}, \textit{testB}), narrowing the gap between supervised and weakly supervised paradigms. 
Under the standard YOLOv3 backbone setting used by most WREC works, compared to DViN, the strongest WREC method, the proposed \nn~consistently outperforms or achieves comparable results across all splits of three datasets, particularly showing significant gains of +3.39\% on RefCOCO \textit{val}, +6.61\% on RefCOCO \textit{testB}, +2.70\% on RefCOCO+ \textit{testA}, and +1.95\% on RefCOCOg \textit{val}. The proposed \nn~surpasses WeakMCN, a strong multi-task learning method, by notable margins of +2.72\% and +2.93\% on RefCOCO \textit{testA} and \textit{testB}, +2.27\% and +2.89\% on RefCOCO+ \textit{val} and \textit{testA}, and +2.37\% on RefCOCOg \textit{val}. These gains confirm the effectiveness and generalization ability of the proposed \nn. We additionally report results using a stronger YOLOv5 backbone. \nn~achieves substantial performance gains, reaching 74.51\% and 76.93\% on RefCOCO \textit{val} and \textit{testA}, respectively. This further validates the robustness of \nn~under modern detection architectures.

\begin{table*}[t]
\caption{Ablation studies on the proposed method. Align.: alignment type. UOR: Unary Object Representation. PRR: Pairwise Relational Representation. FA: flat alignment. CA: compositional alignment.  $\Delta$: performance gain of the proposed method compared to the baseline and WeakMCN ($*$: with ViT-tiny-SAM, $\dagger$: with ViT-small-SAM).
}
\centering
\resizebox{0.9\linewidth}{!}
{
\begin{tabular}{lcccccccc}
\toprule
\small
\multirow{2}{*}{Methods} &\multirow{2}{*}{Align.} & \multicolumn{3}{c}{RefCOCO}                                        & \multicolumn{3}{c}{RefCOCO+}                                       & RefCOCOg                 \\
 & &val& testA  & testB  & val   & testA    & testB     & val       \\ \midrule
Baseline (DINO + Depth) & FA &66.59&69.10&59.94&51.22&58.28&41.62&51.08\\
+ UOR &FA &69.93&71.52&64.30&53.50&58.52&43.08&53.31\\
+ PRR&CA&69.22&71.98&63.47&52.45&58.89&41.97&54.31\\
+ UOR + PRR &CA&\textbf{71.06}&\textbf{72.60}&\textbf{65.55} & \textbf{54.17}&\textbf{60.22}&\textbf{44.63}&\textbf{56.99}\\ 
$\Delta$ &-&$\uparrow4.47$&$\uparrow3.50$&$\uparrow5.61$&$\uparrow2.95$&$\uparrow1.94$&$\uparrow3.01$&$\uparrow5.91$\\\midrule
Baseline (DINO + CLIP) & FA & 63.71&	65.25&	56.09&	48.98	&54.16&	40.15	&50.53\\
+ UOR &FA& 66.61&	68.52	&59.16&	49.68&	56.64&	40.48	&51.12\\
+ PRR &CA& 66.15	&69.33&	58.29&	50.45&	55.64	&40.60	&52.53\\
+ UOR + PRR  & CA  &\textbf{68.90}	&\textbf{70.14}	&\textbf{60.43}	&\textbf{52.77}&\textbf{58.71}&\textbf{42.46}&			\textbf{53.82}\\ 
$\Delta$ &-& $\uparrow5.19$ & $\uparrow4.89$ & $\uparrow4.34$ &$\uparrow3.79$&$\uparrow4.55$&$\uparrow2.31$& $\uparrow3.29$\\

\midrule
WeakMCN$^*$ (CVPR25)~\cite{cheng2025weakmcn} & FA & 68.55 &70.78 &62.00 &51.48&56.92&41.75& 53.44\\
+ UOR + PRR  & CA & \textbf{71.12}& \textbf{72.16} & \textbf{66.10} &\textbf{53.79}&\textbf{59.12}&\textbf{43.94}& \textbf{55.92}\\
$\Delta$ &-&$\uparrow2.57$ & $\uparrow1.38$ & $\uparrow4.10$ &$\uparrow2.31$&$\uparrow2.20$&$\uparrow2.19$&$\uparrow2.48$\\
\midrule
WeakMCN$^\dagger$ (CVPR25)~\cite{cheng2025weakmcn} & FA &69.20 &69.88 &62.63 &51.90& 57.33& 43.10& 54.62 \\
+ UOR + PRR &CA&\textbf{70.54}&\textbf{71.43}&\textbf{64.89}&\textbf{53.23}&\textbf{59.57}&\textbf{44.04}&\textbf{55.94} \\
$\Delta$ &-&$\uparrow1.34$&$\uparrow1.55$&$\uparrow2.26$&$\uparrow1.33$&$\uparrow2.24$&$\uparrow0.96$& $\uparrow1.32$\\
\bottomrule
\end{tabular}
}

\label{tab:abla}
\vspace{-1em}
\end{table*}

\subsection{Quantitative Ablation Analysis}

\noindent\textbf{Effect of Unary Object Representations (UOR) and Pairwise Relational Representations (PRR).} As shown in Table~\ref{tab:abla}, adding either UOR or the PRR consistently improves performance over the baselines under multiple backbone settings (DINO+Depth and DINO+CLIP). On RefCOCO, where expressions are often short, UOR brings strong improvements (\eg, +3.34\% on \textit{val} and +4.36\% on \textit{testB}).  In contrast, on RefCOCOg, which contains long and unconstrained descriptions, PRR is more beneficial (+3.23\% gain on \textit{val}), reflecting the importance of capturing inter-object relations in free-form language. On RefCOCO+, where referring expressions are longer and less position-driven, single modules bring only modest gains, but combining them yields larger improvements. Similar trends are observed across RefCOCO and RefCOCOg, with the largest boost (+5.91\%) achieved when both modules are combined on RefCOCOg. These results suggest that the proposed UOR and PRR play complementary roles, and their synergy is particularly important for handling longer, composition-heavy expressions.

\vspace{3pt}
\noindent\textbf{Compositional Alignment~\textit{v.s.}~Flat Alignment.} Table~\ref{tab:abla} also shows that when the proposed two modules, \ie, Unary Object Representation and Pairwise Relational Representation, are combined under Compositional Alignment (CA), the gains are further improved. CA achieves the best results across all datasets, outperforming Flat Alignment (FA) by a substantial margin (up to +5.91\% on the \textit{val} set of RefCOCOg). Importantly, this trend holds even when integrating our modules into WeakMCN~\cite{cheng2025weakmcn}, a recent multi-task method: CA consistently yields higher accuracy than FA. These results demonstrate that the proposed method provides a general and complementary improvement, enabling effective grounding performances.

\vspace{3pt}
\noindent\textbf{Baseline Analysis.}
Motivated by DViN~\cite{chen2025dvin} and WeakMCN~\cite{cheng2025weakmcn}, we enrich the original anchor features with additional representations extracted from visual foundation models via a dynamic routing mechanism. By default, we use DINOv2 features, which have been widely demonstrated to be semantically rich. We also introduce DepthAnything features to provide depth cues. Table~\ref{tab:abla} shows that our proposed modules consistently bring significant gains across different baselines, \eg, +5.61\% and +4.34\% with ``DINO+Depth'' and ``DINO+CLIP'', respectively, on RefCOCO \textit{testB}. Comparing two baselines, ``DINO+Depth'' is consistently stronger than ``DINO+CLIP''. This suggests that geometric depth cues provide more complementary information to DINO's semantic features than CLIP's visual embeddings, which are semantically aligned but less effective in capturing fine-grained structural distinctions under weak supervision.

\begin{table*}[t]
\caption{Comparison of relational representation forms in WREC. URR: Unary Relational Representations, where relational context is absorbed into anchor embeddings. PRR: Pairwise Relational Representations, where inter-object relations are maintained as independent representation units. UOR: Unary Object Representation.
}\vspace{-1em}
\centering
\resizebox{0.9\linewidth}{!}
{
\begin{tabular}{lcccccccc}
\toprule
\multirow{2}{*}{Methods}&Relational  & \multicolumn{3}{c}{RefCOCO}                                        & \multicolumn{3}{c}{RefCOCO+}                                       & RefCOCOg                 \\
&Representation&  val& testA  & testB     & val& testA    & testB         & val       \\ \midrule
Baseline&None &  66.59&	69.10&	59.94&	51.22	&58.28	&41.62	&51.08 \\
+ URR & Unary  &  \textbf{69.69}	&71.70	&63.20&\textbf{52.83}&58.33&41.19&53.04\\
+ PRR & Pairwise  &  69.22&	\textbf{71.98}&	\textbf{63.47}	&52.45	&\textbf{58.89}	&\textbf{41.97}&	\textbf{54.31}\\
\midrule
+ UOR & None &  69.93&	71.52	&64.30	&53.50	&58.52	&43.08	&53.31\\
+ UOR + URR & Unary & \textbf{71.59}&	72.44	&65.34&	53.01&	58.61	&43.12&	55.11\\
+ UOR + PRR & Pairwise & 71.06	&\textbf{72.60}	&\textbf{65.55}&	\textbf{54.17}	&\textbf{60.22}&	\textbf{44.63}	&\textbf{56.99}\\
\bottomrule
\end{tabular}}
\label{tab:relation}\vspace{-2em}
\end{table*}

\noindent\textbf{Relational Representation Forms.} Table~\ref{tab:relation} compares different relational representation forms in WREC. Both URR (relations absorbed into unary embeddings) and PRR (explicit pairwise representations) improve over the baseline, confirming the benefit of incorporating relational context. On RefCOCO, URR and PRR achieve comparable performance with small variations across splits. This is likely because RefCOCO expressions are generally shorter and less compositionally complex, where unary representations are often sufficient to capture the key grounding cues. In contrast, PRR consistently yields larger gains on composition-heavy settings. For example, on RefCOCOg, PRR improves the baseline by +3.23\%, compared to +1.96\% from URR, indicating that explicitly maintaining pairwise relations better captures structured inter-object dependencies.
When combined with UOR, the advantage of PRR becomes more evident. On RefCOCOg, adding PRR on top of UOR yields a +3.68\% gain over UOR alone, clearly higher than the +1.80\% improvement from URR. 
These results suggest that while relational aggregation is helpful, flattening relations into unary embeddings is insufficient for modeling structured compositional semantics. Explicit pairwise representations provide a more effective and robust solution for weakly supervised grounding.

\begin{figure*}[t]
    \centering
    \includegraphics[width=0.95\linewidth]{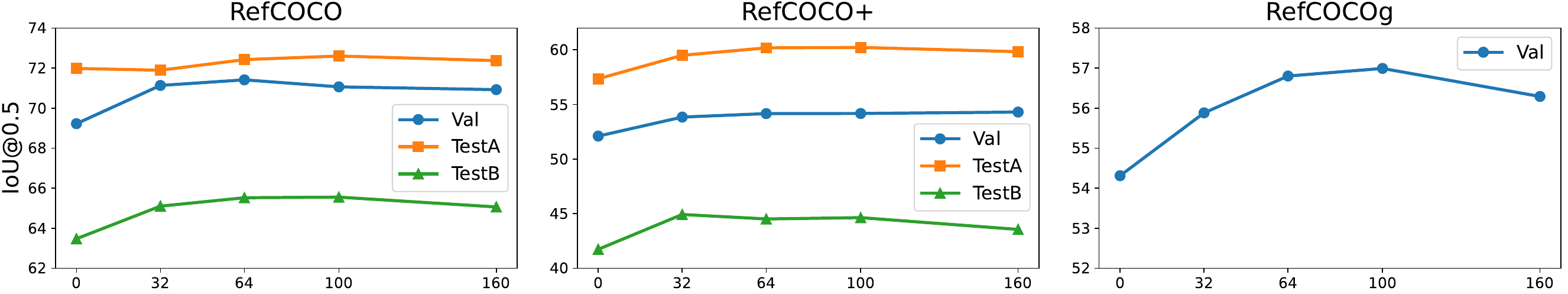}
    \caption{Effect of varying the number of semantic basis vectors.
    }
    \label{fig:attributes}
    \vspace{-2em}
\end{figure*}

\noindent\textbf{Effect of the Number of Semantic Basis Vectors.}
Fig.~\ref{fig:attributes} shows the effect of varying the number of semantic basis vectors in Unary Object Representation. The performance generally improves as the number increases across all datasets up to 100, suggesting that a richer basis set provides greater semantic capacity for representing object characteristics. The performance declines on most datasets when learning more than 100 basis vectors. This is likely due to redundancy and overfitting. Therefore, a moderate basis size (\eg, 100, as used in our main experiments) provides the best balance between diversity and distinctiveness.

\begin{wraptable}{r}{0.48\textwidth}
\vspace{-3em}
\caption{Effect of $\mathcal{L}_{\text{div}}$ and $\mathcal{L}_{\text{cons}}$. }
\label{abla_loss}
\centering
\resizebox{0.9\linewidth}{!}{

\begin{tabular}{cccccc}
\toprule
\multirow{2}{*}{$\mathcal{L}_{\text{div}}$}&\multirow{2}{*}{$\mathcal{L}_{\text{cons}}$} & \multicolumn{2}{c}{RefCOCO}                                        & \multicolumn{2}{c}{RefCOCO+}   \\
&&testA  & testB     & testA    & testB       \\ \midrule
&&71.88	&64.59 &58.47 &43.58\\
\checkmark & &71.75&	65.48& 59.27	&43.51\\
&\checkmark &72.21	&64.46& 59.48	&44.39\\
\checkmark & \checkmark &\textbf{72.60}	&\textbf{65.55}& \textbf{60.22}&	\textbf{44.63}\\
\bottomrule 
\end{tabular}
}
\vspace{-1.5em}
\end{wraptable}

\noindent\textbf{Effect of Different Regularization Losses.}
As shown in Table~\ref{abla_loss}, applying diversity loss ($\mathcal{L}_{\text{div}}$) alone improves performance from 64.59\% to 65.48\% on RefCOCO \textit{testB}, which contains more non-human related expressions compared to other splits that are dominated by human-centric descriptions. This indicates that promoting diverse semantic basis vectors is especially beneficial for handling broader and less familiar categories. In contrast, the consistency loss ($\mathcal{L}_{\text{cons}}$), which enforces alignment consistency between unary-chunk and pairwise–sentence matching, is particularly effective on RefCOCO+ (\textit{i.e.}, from 48.47\% to 59.48\% on \textit{testA} and from 43.58\% to 44.39\% on \textit{testB}), where expressions are longer and rich in relational semantics. Combining these two regularizations achieves the best results across all datasets. This demonstrates their effectiveness in strengthening the generalization ability of the framework across diverse linguistic scenarios.

\begin{figure*}[t]
    \centering
    \includegraphics[width=0.95\linewidth]{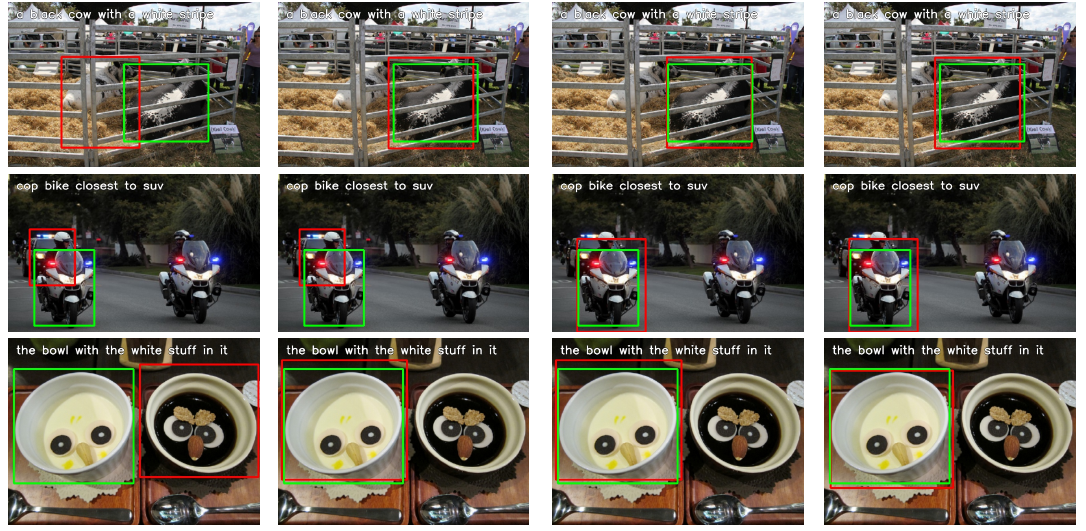}
    \caption{Qualitative results of the proposed method under different configurations. From left to right, each column corresponds to the baseline, baseline with Unary Object Representation, baseline with Pairwise Relational Representation, and \nn, respectively. Red boxes: predictions. Green boxes: GT. See more visualizations in Supplementary. \label{fig:vis}
    }
\end{figure*}

\subsection{Qualitative Analysis}
\noindent\textbf{Component-wise Analysis.} Fig.~\ref{fig:vis} shows how the proposed modules, \ie, Unary Object Representation (UOR) and Pairwise Relational Representation (PRR), contribute to grounding performance under different types of referring expressions. For expressions that primarily describe the object itself, UOR significantly improves localization accuracy by enlarging the anchors' instance-level distinctiveness with learned attribute cues, \eg, in the first example ``a black cow with white stripe'', it helps the model focus on the cow with the distinctive stripe. In contrast, when expressions emphasize relations between objects (\eg, the second row, ``cop bike closest suv''), PRR proves highly effective, correctly disambiguating the target ``bike'' among multiple visually similar candidates by explicitly modeling inter-anchor dependencies. The full \nn~model, which integrates both modules, achieves the most precise localization by capturing both attribute-level cues and inter-object relations.

\begin{figure*}[t]
    \centering
    \includegraphics[width=\linewidth]{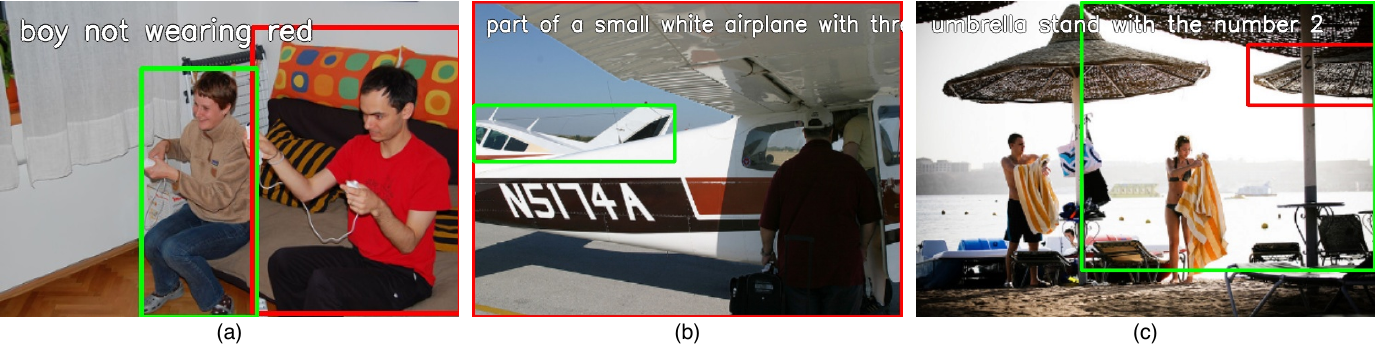}
    \caption{Failure cases. The referring texts are (a) \textit{boy not wearing red}, (b) \textit{part of a small white airplane with three windows}, (c) \textit{umbrella stand with the number 2}. Red boxes: predictions. Green boxes: GT. See more examples and analysis in the Supplementary.\label{fig:failure_cases}
    }
\end{figure*}

\noindent\textbf{Failure Analysis.} Fig~\ref{fig:failure_cases} shows representative failure cases.
First, the model struggles with expressions involving negation, as shown in Fig.~\ref{fig:failure_cases}(a), where the correct target must be identified by excluding objects that satisfy the negated attribute. Addressing such cases may benefit from explicitly modeling negation or exclusion reasoning, which remains an interesting direction for future work.
Second, failures may arise under complex visual ambiguity, as shown in Fig.~\ref{fig:failure_cases}(b-c). In these examples, the target objects are either partially occluded, visually ambiguous, or surrounded by salient distractors with similar appearance. Additional failure cases and more detailed analysis are provided in the Supplementary.

\section{Conclusion}
We proposed Structured Visual Compositional Representations (\nn) for Weakly Supervised Referring Expression Comprehension. Unlike prior approaches that operate on flat anchor-level representations and align individual anchors directly with textual embeddings, our framework explicitly constructs a structured visual representation space composed of complementary unary object embeddings and pairwise relational embeddings.
These two components are aligned with subject-level and sentence-level linguistic cues through a proposed compositional alignment mechanism, enabling the model to jointly capture instance-level attributes and structured inter-object dependencies under weak supervision. 
Extensive experiments on RefCOCO, RefCOCO+, and RefCOCOg demonstrate consistent improvements over existing WREC methods across multiple backbone settings, validating the robustness and generality of the proposed structured modeling paradigm. We believe this structured representation perspective offers a principled direction for advancing weakly supervised vision–language grounding beyond flat anchor-based alignment.

\section*{Acknowledgements}
This research was supported in part by the Australian Research Council (ARC) DECRA Fellowship (DE260101852), ARC Disccovery Project (DP260101891), ARC Future Fellowship (FT250100448), the National Natural Science Foundation of China (No. 62372491), Early Career Scheme of the Research Grants Council (RGC) of the Hong Kong SAR under grant No. 26202321, ITF PRP/046/24FX, Science \& Technology Cooperation Program of Shandong under grant No. SDST26EG01, SAIL Research Project, HKUST-Zeekr Coolaborative Research Fund.

\bibliographystyle{splncs04}
\bibliography{main}

\renewcommand{\thesection}{S\arabic{section}}
\renewcommand{\thefigure}{S\arabic{figure}}
\renewcommand{\thetable}{S\arabic{table}}
\renewcommand{\theequation}{S\arabic{equation}}

\setcounter{section}{0}
\setcounter{figure}{0}
\setcounter{table}{0}
\setcounter{equation}{0}
\input{supp}

\end{document}

%% file: supp.tex
\title{Learning Structured Visual Compositional Representations for Weakly Supervised Referring Expression Comprehension -- Supplementary Materials}




\titlerunning{Learning Structured Visual Compositional Representations for WREC}



\author{\mbox{}}
\institute{\mbox{}}

\maketitle
\vspace{-1.5cm}

\section{Additional Experimental Details}

\vspace{-0.1em}

\noindent\textbf{Dataset Details.} RefCOCO contains 142,210 expressions referring to 50,000 objects in 19,994 images. It contains relatively short and direct expressions, which include more spatial descriptions. 
RefCOCO+ includes 141,564 expressions referring to 49,856 objects in 19,992 images, but explicitly discourages the use of spatial terms during annotation, thus having more appearance-based descriptions.
RefCOCOg incudes 104,560 expressions, which are longer and more complex than those in RefCOCO and RefCOCO+. It covers 54,822 objects in 26,711 images, with more compositional descriptions about both appearance and spatial information.

\noindent\textbf{Implementation Details.} 
Following prior work~\cite{jin2023refclip, chen2025dvin, cheng2025weakmcn}, we used the pretrained YOLOv3 as the default detection network unless otherwise specified, and set the number of candidate anchors $K$ to 17. 
Following DViN~\cite{chen2025dvin} and WeakMCN~\cite{cheng2025weakmcn}, we incorporated pretrained visual foundation models to enrich anchor features. More specifically, we used DINOv2-base~\cite{oquab2024dinov2} model. For baseline comparisons, we used the CLIP ViT-base~\cite{radford2021learning} vision encoder. We additionally used the Depth Anythingv2-small~\cite{yangdepth} model. We initialized the text embeddings with GloVe~\cite{pennington2014glove} pretrained weights and set the maximum sequence length to 15. We used the SpaCy python library to extract the subject noun chunk in each referring expression. The training image size was set to $416\times 416$. The training used a 3-epoch warm-up step, with the learning rate linearly increasing from 1e-4 to 1e-3. After the warm-up, the learning rate was set at 1e-3 with a cosine decay schedule. The model is trained for a total of 25 epochs. 
All experiments were conducted on a single NVIDIA RTX 3090 GPU.

\begin{table*}[t]
\caption{Performance comparison with the state-of-the-art WREC methods on RefCOCO, RefCOCO+, and RefCOCOg. $^*$: with ViT-tiny-SAM, $^\dagger$: with ViT-small-SAM. $^{^\ddagger}$ denotes results based on YOLOv5, others are based on YOLOv3.
The best and the second best results are \textbf{bold} and \underline{underlined}.
}
\vspace{-0.5em}
\centering
\resizebox{\linewidth}{!}
{
\begin{tabular}{lccccccccg}
\toprule
\multirow{2}{*}{Methods} & \multicolumn{3}{c}{RefCOCO}                                        & \multicolumn{3}{c}{RefCOCO+}                                       & RefCOCOg           & \cellcolor[gray]{0.9} Inference      \\
& val& testA  & testB     & val& testA    & testB         & val     & \cellcolor[gray]{0.9} Speed  \\ \midrule
WeakMCN$^*$ (CVPR25)~\cite{cheng2025weakmcn} & 68.55 &70.78 &62.00 &51.48&56.92&41.75& 53.44 & \cellcolor[gray]{0.9} 17.27 FPS\\
WeakMCN$^\dagger$ (CVPR25)~\cite{cheng2025weakmcn}  &69.20 &69.88 &62.63 &51.90& 57.33& 43.10& 54.62 & \cellcolor[gray]{0.9} 16.92 FPS\\
DViN (CVPR25)~\cite{chen2025dvin}  &  67.67& 70.90& 59.39 &52.54& 57.52 &\underline{45.31} &55.04 & \cellcolor[gray]{0.9} 14.49 FPS\\
\nn~(Ours) & \underline{71.06}&\underline{72.60}&\underline{65.55} & \underline{54.17}&\underline{60.22}&44.63&\underline{56.99} & \cellcolor[gray]{0.9} 18.28 FPS\\
\nn$^{^\ddagger}$ (Ours) & \textbf{74.51}&	\textbf{76.93}	&\textbf{70.46}&	\textbf{56.44}&	\textbf{64.15}	&\textbf{48.68}&	\textbf{60.40} & \cellcolor[gray]{0.9} \textbf{20.20 FPS}\\
\bottomrule
\end{tabular}}
\label{tab:sota_FPS}
\end{table*}

\begin{table*}[t]
\caption{Ablation studies on the proposed method. Align.: alignment type. UOR: Unary Object Representation. PRR: Pairwise Relational Representation. FA: flat alignment. CA: compositional alignment.  $\Delta$: performance gain of the proposed method compared to the baseline.
}
\vspace{-0.5em}
\centering
\resizebox{\linewidth}{!}
{
\begin{tabular}{lccccccccc}
\toprule
\small
\multirow{2}{*}{Methods} &\multirow{2}{*}{Align.} & \multicolumn{3}{c}{RefCOCO}                                        & \multicolumn{3}{c}{RefCOCO+}                                       & RefCOCOg & \cellcolor[gray]{0.9} Inference                \\
 & &val& testA  & testB  & val   & testA    & testB     & val    & \cellcolor[gray]{0.9} Speed   \\ \midrule
Baseline (DINO + Depth) & FA &66.59&69.10&59.94&51.22&58.28&41.62&51.08 & \cellcolor[gray]{0.9} 19.50 FPS\\
+ UOR &FA &69.93&71.52&64.30&53.50&58.52&43.08&53.31 & \cellcolor[gray]{0.9} 19.11 FPS \\
+ PRR&CA&69.22&71.98&63.47&52.45&58.89&41.97&54.31 & \cellcolor[gray]{0.9} 18.79 FPS\\
+ UOR + PRR (\nn)&CA&\textbf{71.06}&\textbf{72.60}&\textbf{65.55} & \textbf{54.17}&\textbf{60.22}&\textbf{44.63}&\textbf{56.99} & \cellcolor[gray]{0.9} 18.28 FPS\\ 
$\Delta$ &-&$\uparrow4.47$&$\uparrow3.50$&$\uparrow5.61$&$\uparrow2.95$&$\uparrow1.94$&$\uparrow3.01$&$\uparrow5.91$&\cellcolor[gray]{0.9} $\downarrow1.22$ FPS\\

\bottomrule
\end{tabular}
}

\label{tab:abla_FPS}

\end{table*}

\section{Additional Quantitative Results.} \noindent\textbf{Inference Efficiency.} We further evaluate the inference efficiency of our method and report the results in Table~\ref{tab:sota_FPS} and Table~\ref{tab:abla_FPS}.
To ensure fair comparisons, all inference speed measurements are obtained under a unified evaluation protocol. More specifically, experiments are conducted on a single NVIDIA RTX 3090 GPU with batch size = 1 and an input resolution of 416 × 416. The throughput is measured on the RefCOCO validation split using CUDA event timing, with a warm-up phase of 30 iterations. The reported FPS (Frames Per Second) corresponds to the average forward-pass throughput over the entire dataset.

Since most existing WREC methods adopt YOLOv3 as the visual backbone, we first report results using a YOLOv3-based implementation for fair comparison. As shown in Table~\ref{tab:sota_FPS}, our method achieves 18.28 FPS, which is comparable to or faster than prior methods while delivering substantially higher grounding accuracy.
The inference speed differences mainly arise from the feature extraction pipeline. DViN~\cite{chen2025dvin} adopts four foundation models (DINOv2, SAM, CLIP-ViT, and CLIP-CNN), and WeakMCN~\cite{cheng2025weakmcn} adopts a multi-task architecture with additional features from DINOv2 and SAM. In contrast, our method only uses DINOv2 and the DepthAnything-v2-small model. The proposed modules operate on a small set of anchor features and consist of lightweight operations, resulting in negligible additional inference cost.
We additionally evaluate our framework with a stronger YOLOv5 backbone. The results show a clear improvement in grounding performance while the inference speed further increases to 20.20 FPS. This suggests that future WREC methods could benefit from adopting more advanced detection backbones.
Table~\ref{tab:abla_FPS} further analyzes the computational overhead of different components. Compared to the baseline (19.50 FPS), introducing unary object representations (UOR) and pairwise relational representations (PRR) leads to only a modest speed reduction, with the full model running at 18.28 FPS, corresponding to a 1.22 FPS decrease. This demonstrates that the proposed components introduce minimal additional computational overhead while providing consistent performance improvements, indicating that our approach achieves a favorable accuracy–efficiency trade-off.

\begin{figure}[t]
    \centering
    
    \includegraphics[width=0.75\linewidth]{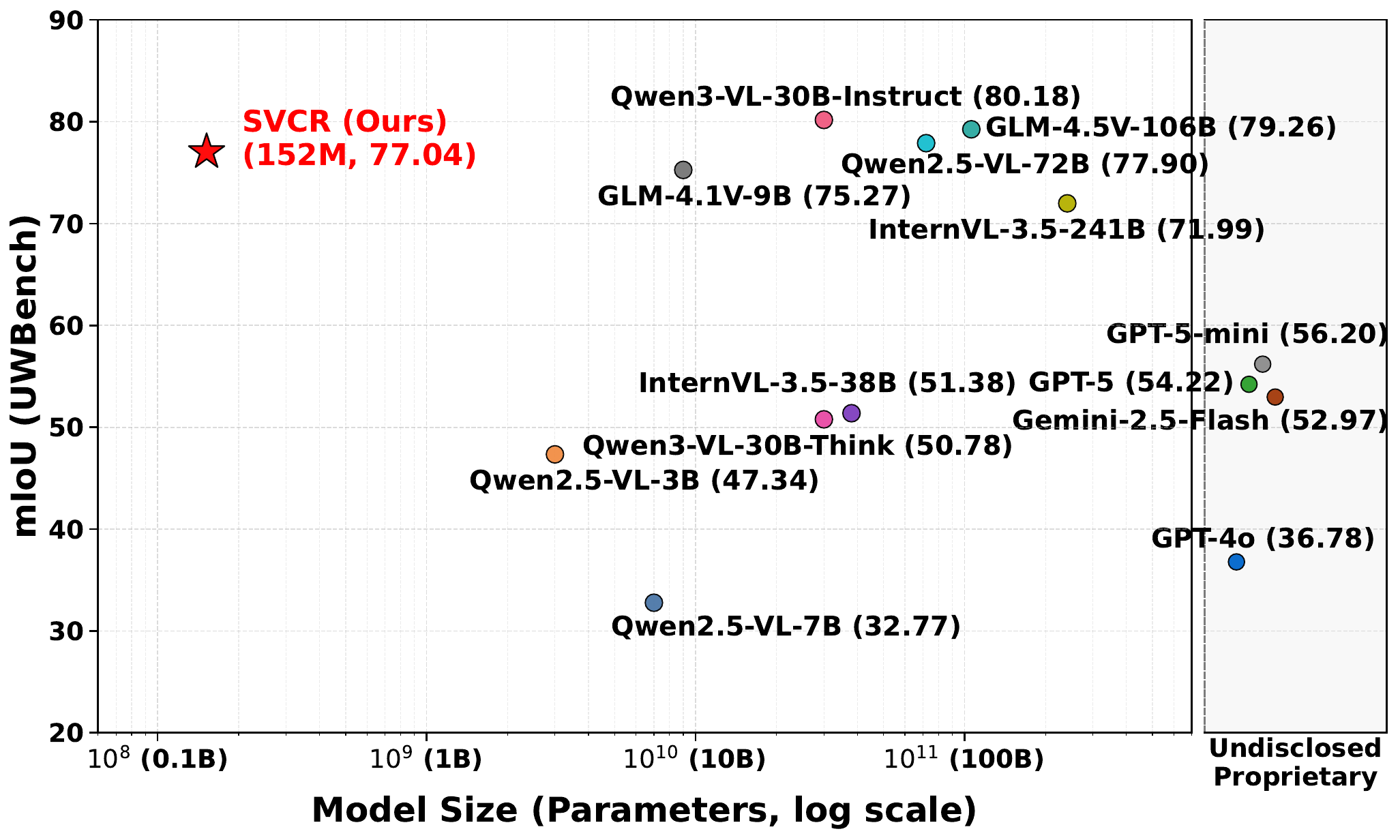}
    \vspace{-1em}
    \caption{Accuracy-efficiency comparison on UWBench.
    }
    \label{fig:uwbench}
\end{figure}

\begin{table}[t]
\vspace{-1em}
\caption{Results using YOLOv5. ($*$: with ViT-tiny-SAM, $\dagger$: with ViT-small-SAM)
}
\centering
\begin{tabular}{lccccccc}   
\toprule
\multirow{2}{*}{Methods}& \multicolumn{3}{c}{RefCOCO}                                        & \multicolumn{3}{c}{RefCOCO+}                                       & RefCOCOg    \\
& val& testA  & testB     & val& testA    & testB         & val     \\ \midrule
WeakMCN$^*$[4] &72.41&74.19&68.79& 54.57 & 63.34 & 46.98 &57.88\\
WeakMCN$^\dagger$[4] &71.66 & 73.11& 68.75 & 53.27& 61.67 & 45.08 & 56.82\\
SVCR (Ours) & \textbf{74.51}&	\textbf{76.93}	&\textbf{70.46}&	\textbf{56.44}&	\textbf{64.15}	&\textbf{48.68}&	\textbf{60.40} \\
\bottomrule
\end{tabular}

\label{tab:yolov5}
\vspace{-1em}
\end{table}

\noindent\textbf{Comparison under YOLOv5.} Table~\ref{tab:yolov5} compares SVCR with WeakMCN variants under the YOLOv5 setting. Compared with the YOLOv3 results reported in the main paper, WeakMCN also shows clear improvements with the stronger backbone, suggesting that better candidate generation generally benefits anchor-based WREC methods. Under the same YOLOv5 setting, SVCR consistently outperforms WeakMCN across all splits of RefCOCO, RefCOCO+, and RefCOCOg. SVCR, like other anchor-based WREC methods~\cite{jin2023refclip,luo2024apl,chen2025dvin,cheng2025weakmcn}, depends on the quality of candidate anchors. If the detector fails to cover the target object, the downstream modules cannot recover the correct grounding result. This is a shared limitation of the anchor-based WREC paradigm.

\noindent\textbf{Additional Results on UWBench.} We additionally evaluated SVCR on UWBench~\cite{zhang2025uwbench}, a recent underwater vision-language benchmark containing covering 158 underwater categories. Unlike RefCOCO, UWBench involves underwater-specific visual degradation, marine taxonomy, morphological attributes, and spatial relations. 
We trained SVCR on UWBench using only image–referring text pairs. Fig.~\ref{fig:uwbench} shows that SVCR outperforms several large general VLMs reported in~\cite{zhang2025uwbench}, including GPT-5, Gemini-2.5-Flash, InternVL-3.5-24B, GLM-4.1V-9B, and Qwen2.5-VL-7B, and it performs competitively with much larger models such as Qwen2.5-VL-72B and GLM-4.5V-106B, while using approximately 470×–700× fewer parameters. This accuracy–efficiency trade-off highlights a key advantage of WREC: rather than relying on costly cloud-scale VLMs, a compact weakly supervised model can provide strong target-domain grounding with substantially lower computational cost. We further evaluated 32/64/100/160 bases on UWBench, obtaining 76.56/77.11/77.04/76.72 mIoU. The 100-base setting is within 0.07\% of the best, confirming robustness without dataset-specific tuning.

\begin{table*}[t]
\caption{Effect of the number of selected anchors ($K$).
}
\centering
\begin{tabular}{lccccccccc}
\toprule
\multirow{2}{*}{Methods} &\multirow{1}{*}{\#Anchors} & \multicolumn{3}{c}{RefCOCO}                                        & \multicolumn{3}{c}{RefCOCO+}                                       & RefCOCOg    \\
&($K$)& val& testA  & testB     & val& testA    & testB         & val     \\ \midrule
\nn & 17 & 71.06	&\textbf{72.60}&	65.55	&\textbf{54.17}	&\textbf{60.22}&	44.63	&\textbf{56.99} \\
\nn & 34 & \textbf{72.51}	&71.70	&\textbf{66.36}	&54.02	&60.11	&\textbf{46.12}	&53.08 \\
\nn & 68 & 71.19	&71.35&	66.30&	50.15	&55.12	&43.98&	46.43 \\
\bottomrule
\end{tabular}
\label{tab:select_num}
\end{table*}

\noindent\textbf{Effect of the number of selected anchors.}
Following prior WREC methods~\cite{luo2024apl,cheng2025weakmcn,chen2025dvin}, we adopt ($K$=17) anchors in all experiments. We additionally analyze the impact of the number of selected anchors in Table~\ref{tab:select_num}. Increasing $K$ from 17 to 34 improves performance on the RefCOCO \textit{val} and \textit{testB} splits, and the RefCOCO+ \textit{testB} split, while reducing the performance on the RefCOCOg \textit{val} set. When $K$ is further increased to 68, the performance consistently drops across three datasets, particularly with significant degradation on RefCOCO+ and RefCOCOg. 
On RefCOCO and RefCOCO+, where expressions are typically shorter and rely on relatively coarse attributes or spatial cues, moderately increasing the number of anchors can provide richer candidate regions and improve the possibility of covering the target. In contrast, RefCOCOg contains longer and more descriptive expressions that require more precise semantic alignment. Introducing too many anchors in this case increases the number of visually similar distractors, which can weaken the discriminative matching among anchors under weak supervision and make the correct target harder to identify. These results suggest that a moderate anchor set provides a better balance between candidate coverage and distractor suppression, and the commonly adopted setting ($K$=17) remains a robust choice across datasets.

\noindent\textbf{Parsing Error Analysis.}
To obtain the subject phrase used for unary alignment, we parse each referring expression using spaCy (\texttt{en\_core\_web\_lg}) and extract noun phrases via \texttt{doc.noun\_chunks}. We then select the \emph{first} noun phrase (NP) as the subject chunk. If no NP is detected, the full expression is used. This heuristic is motivated by the linguistic prior in referring expressions that the referent object typically appears at the beginning of the sentence (\eg, ``the tall man''), while the remaining clauses often describe relations (\eg, ``wearing a red hat'', ``next to the car'').
To assess potential parsing errors, we analyze the RefCOCOg validation set, which contains longer and structurally richer expressions. In this set, 85.9\% of expressions contain multiple noun chunks. As a reference, we employ an LLM (GPT-4.1-mini) to extract the referent subject phrase. Each referring expression is provided with instructions to identify the subject noun phrase corresponding to the referent object while ignoring relational clauses or attributes describing other objects. The model is required to return the extracted phrase in a structured output format to ensure consistent parsing.
We compare spaCy’s first-NP extraction with the oracle output using both exact match and a relaxed token-level F1 metric. The two methods agree in 75.5\% exact match and 77.2\% relaxed match, with an average token-level F1 of 0.886. The agreement is higher for relational (multi-NP) expressions (77.0\% exact, F1 = 0.892), suggesting that the leading noun phrase typically identifies the referent.
These results suggest that subject NP extraction is generally reliable and unlikely to be a bottleneck. Moreover, the pipeline is parser-agnostic, namely stronger extractors, including LLM-based methods, can be readily incorporated to handle more complex cases if desired.

\begin{figure*}[t]
    \centering
    \includegraphics[width=\linewidth]{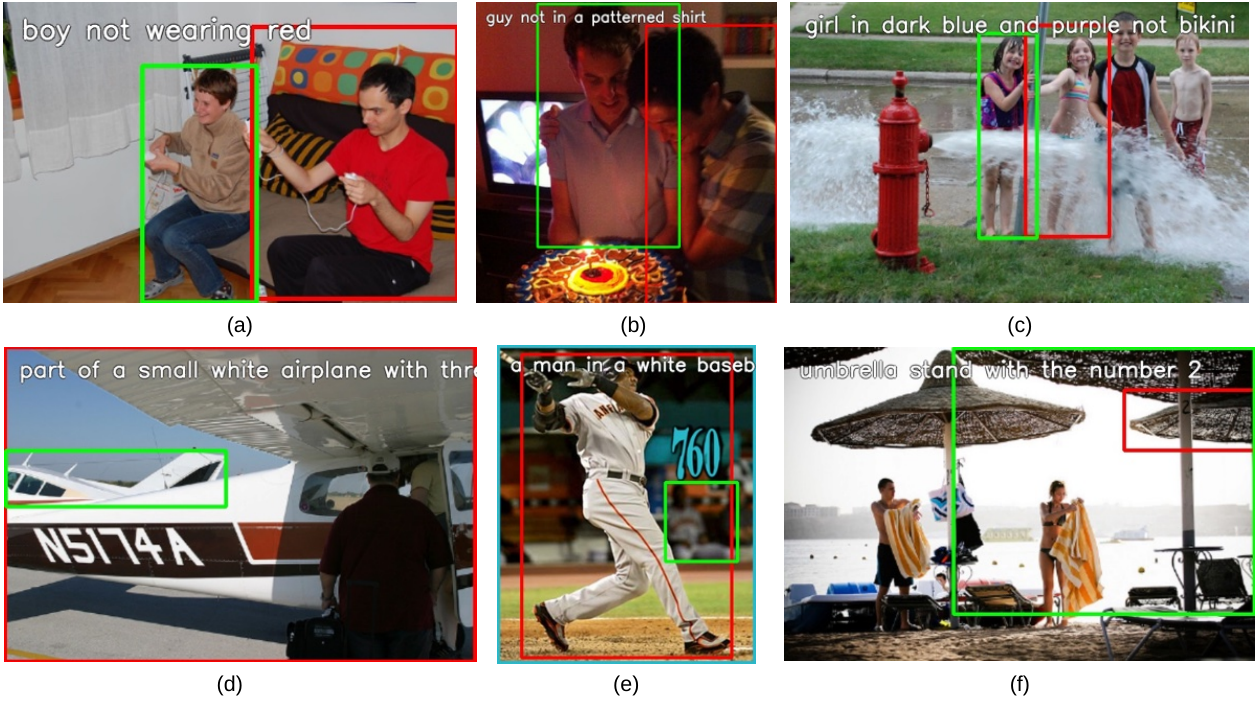}
    \caption{Failure cases. The referring texts are (a) \textit{boy not wearing red}, (b) \textit{guy not in a patterned shirt}, (c) \textit{girl in dark blue and purple not bikini}, (d) \textit{part of a small white airplane with three windows}, (e) \textit{a man in a white baseball uniform waiting in the dug out}, (f) \textit{umbrella stand with the number 2}. Red boxes: predictions. Green boxes: GT. \label{fig:failure_cases_supp}
    }
\end{figure*}

\section{Additional Qualitative Results.} 
\noindent\textbf{Analysis of Failure Cases.} Fig.~\ref{fig:failure_cases_supp} presents several failure cases that highlight two common challenges in WREC.
First, the model struggles with expressions involving negation, as shown in Fig.~\ref{fig:failure_cases_supp}(a–c), where the correct target must be identified by excluding objects that satisfy the negated attribute. Addressing such cases may benefit from explicitly modeling negation or exclusion reasoning, which remains an interesting direction for future work.
Second, failures may arise under complex visual ambiguity, as shown in Fig.~\ref{fig:failure_cases_supp}(d–f). In these examples, the target objects are either partially occluded, visually ambiguous, or surrounded by salient distractors with similar appearance. For instance, in Fig.~\ref{fig:failure_cases_supp}(d), the large foreground airplane body dominates the scene, while the target region corresponds only to a small visible part of another airplane with three windows. The strong visual saliency of the foreground airplane and their shared white appearance make the target difficult to distinguish. Similarly, in Fig.~\ref{fig:failure_cases_supp}(e), the large baseball player in the foreground visually resembles the target person in the dugout, as both are men wearing white baseball uniforms. The strong foreground presence and the blurred background appearance lead the model to focus on the prominent foreground subject instead of the intended background target. In Fig.~\ref{fig:failure_cases_supp}(f), the target pole overlaps with another umbrella structure and is specified only by a small visual cue (the number ``2''). These cases illustrate that grounding errors mainly occur under challenging visual conditions or subtle linguistic constraints that go beyond the primary focus of the proposed method.

\begin{figure*}[ht]
    \centering
    \includegraphics[width=\linewidth]{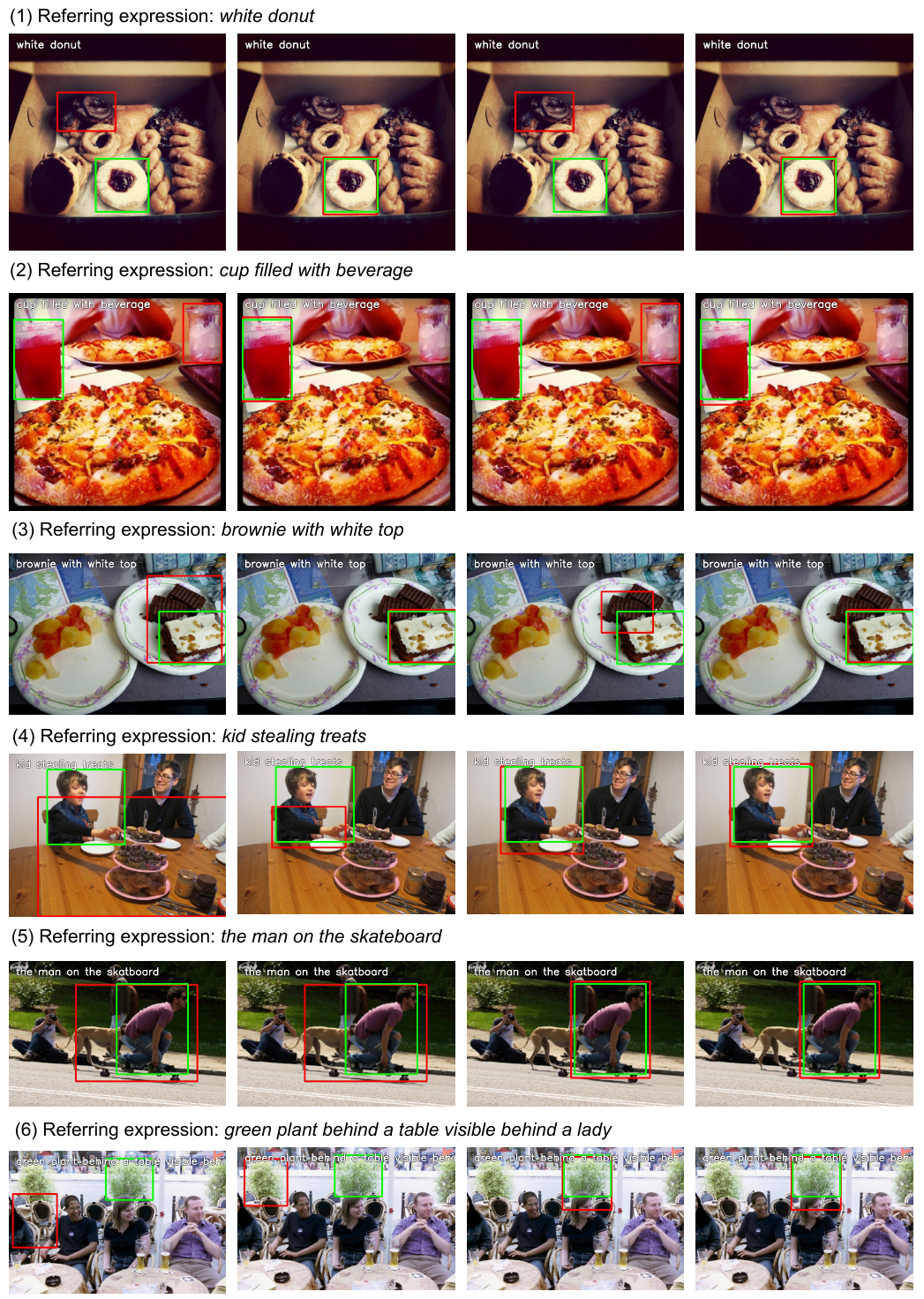}
    \caption{Qualitative results. From left to right, each column corresponds to the baseline, baseline with Unary Object Representations (UOR), baseline with Pairwise Relational Representations (URR), and our full model \nn~(with both UOR and URR), respectively. Red boxes: predictions. Green boxes: GT. \label{fig:vis_appendix}
    }
\end{figure*}

\noindent\textbf{Qualitative Analysis of Model Components.} Fig.~\ref{fig:vis_appendix} provides additional qualitative examples to illustrate how the proposed method improves grounding under weak supervision. For attribute-focused descriptions (\eg, ``white donut'', ``cup filled with beverage'', ``brownie with white top''), adding Unary Object Representations (UOR) consistently improves localization by enabling anchors to be more discriminative, whereas the baseline often confuses the target with visually similar instances. For relation-based expressions (\eg, ``the man on the skateboard'' and the action-oriented ``kid stealing treats''), 
Pairwise Relational Representations (PRR) better captures inter-object dependencies and directional relations, enabling accurate localization where the baseline or unary-only variant tends to mis-cover nearby or co-occurring objects. In a more complex case of ``green plant behind a table visible behind a lady'', the baseline mislocalizes the table, and the unary-only variant identifies a plant but not the correct instance. In contrast, the pairwise variant correctly localizes the target plant, as does the full model. These results demonstrate that UOR and PRR capture complementary aspects of referring expressions, and combining them through compositional alignment yields robust grounding across diverse linguistic structures.

\clearpage
